\documentclass[sigconf, screen]{acmart}
\AtBeginDocument{%
  }

\setcopyright{acmlicensed}
\copyrightyear{2026}
\acmYear{2026}
\acmDOI{XXXXXXX.XXXXXXX}
\acmConference[Under review]{Make sure to enter the correct
  conference title from your rights confirmation email}{Arxiv pre-print}{}
\acmISBN{978-1-4503-XXXX-X/2018/06}

\usepackage{thmtools}
\usepackage{enumitem}

\usepackage{algorithm}
\usepackage{algpseudocode}

\declaretheoremstyle[
    headfont=\bfseries,
    notefont=\bfseries,
    bodyfont=\normalfont\itshape,
    headpunct={.},
    postheadspace=0.5em,
    spaceabove=6pt,
    spacebelow=6pt
]{axiomstyle}

\acmSubmissionID{2290}
\renewcommand\footnotetextcopyrightpermission[1]{}
\begin{document}

\title{Open-Loop Planning, Closed-Loop Verification: Speculative Verification for VLA}

\author{Zihua Wang}
\affiliation{%
  \institution{School of Computer Science and Engineering, Southeast University}
  \city{Nanjing}
  \postcode{210096}
  \country{China}
}

\author{Zhitao Lin}
\affiliation{%
  \institution{School of Computer Science and Engineering, Southeast University}
  \city{Nanjing}
  \postcode{210096}
  \country{China}
}

\author{Ruibo Li}
\affiliation{%
  \institution{Nanyang Technological University}
  \country{Singapore}
}

\author{Yu Zhang}
\authornote{Yu Zhang is the corresponding author.}
\affiliation{%
  \institution{School of Computer Science and Engineering, Southeast University}
  \city{Nanjing}
  \postcode{210096}
  \country{China}
}
\email{zhang_yu@seu.edu.cn}

\author{Xu Yang}
\affiliation{%
  \institution{School of Computer Science and Engineering, Southeast University}
  \city{Nanjing}
  \postcode{210096}
  \country{China}
}

\author{Siya Mi}
\affiliation{%
  \institution{School of Cyber Science and Engineering, Southeast University}
  \city{Nanjing}
  \postcode{211189}
  \country{China}
}
\affiliation{%
  \institution{Purple Mountain Laboratories}
  \city{Nanjing}
  \postcode{210000}
  \country{China}
}

\author{Xiu-Shen Wei}
\affiliation{%
  \institution{School of Computer Science and Engineering, Southeast University}
  \city{Nanjing}
  \postcode{210096}
  \country{China}
}

\renewcommand{\shortauthors}{Wang et al.}

\begin{abstract}
Vision-Language-Action (VLA) models, as large foundation models for embodied control, have shown strong performance in manipulation tasks. 
However, their performance comes at high inference cost. 
To improve efficiency, recent methods adopt action chunking, which predicts a sequence of future actions for open-loop execution.
Although effective for reducing computation, open-loop execution is sensitive to  environmental changes and prone to error accumulation due to the lack of close-loop feedback.
To address this limitation, we propose Speculative Verification for VLA Control (SV-VLA), a framework that combines efficient open-loop long-horizon planning with lightweight closed-loop online verification.
Specifically, SV-VLA uses a heavy VLA as a low-frequency macro-planner to generate an action chunk together with a planning context, while a lightweight verifier continuously monitors execution based on the latest observations. 
Conditioned on both the current observation and the planning context, the verifier compares the planned action against a closed-loop reference action and triggers replanning only when necessary.
Experiments demonstrate that SV-VLA combines the efficiency of chunked prediction with the robustness of closed-loop control, enabling efficient and reliable VLA-based control in dynamic environments.
Code is available \href{https://github.com/edsad122/SV-VLA}{here}.
\end{abstract}



\maketitle

\section{Introduction}
\label{sec:intro}

Vision–Language–Action (VLA) models have recently emerged as a promising paradigm in embodied AI, by directly predicting action tokens from multimodal observations and language instructions~\cite{zhang2025pure,jiang2025survey}.
By combining large-scale vision-language pretraining with robot demonstration data, recent VLA models such as RT-2~\cite{zitkovich2023rt}, OpenVLA~\cite{kim2024openvla}, and $\pi_0$~\cite{black2024pi0, intelligence2025pi05,intelligence2025pi} have shown strong zero-shot transfer ability and improved semantic grounding~\cite{yu2025point}.
These properties make them promising for complex, multi-step manipulation tasks and open-world settings, which are difficult to handle with conventional task-specific policies trained on limited robot data~\cite{zhu2025objectvla,xu2024survey}.
However, a key challenge in deploying VLA models lies in their  inference efficiency~\cite{yu2025survey, ma2025running}.

Closed-loop action prediction generates each action conditioned on the latest observation, enabling continuous adaptation to environmental changes~\cite{sendai2025leave,liu2026world}.
However, this sequential decision process introduces significant inference latency, as the model must be executed repeatedly at every timestep. 
To improve efficiency, many recent methods adopt action chunking~\cite{zhao2023learning,oft}, which predicts a short horizon of future actions and executes them in an open-loop manner.
Nonetheless, this action chunking mechanism executes actions based on stale observations and fails to incorporate newly observed feedback during chunk execution, leading to error accumulation and degraded control performance.

\begin{figure*}
  \includegraphics[width=0.96\textwidth]{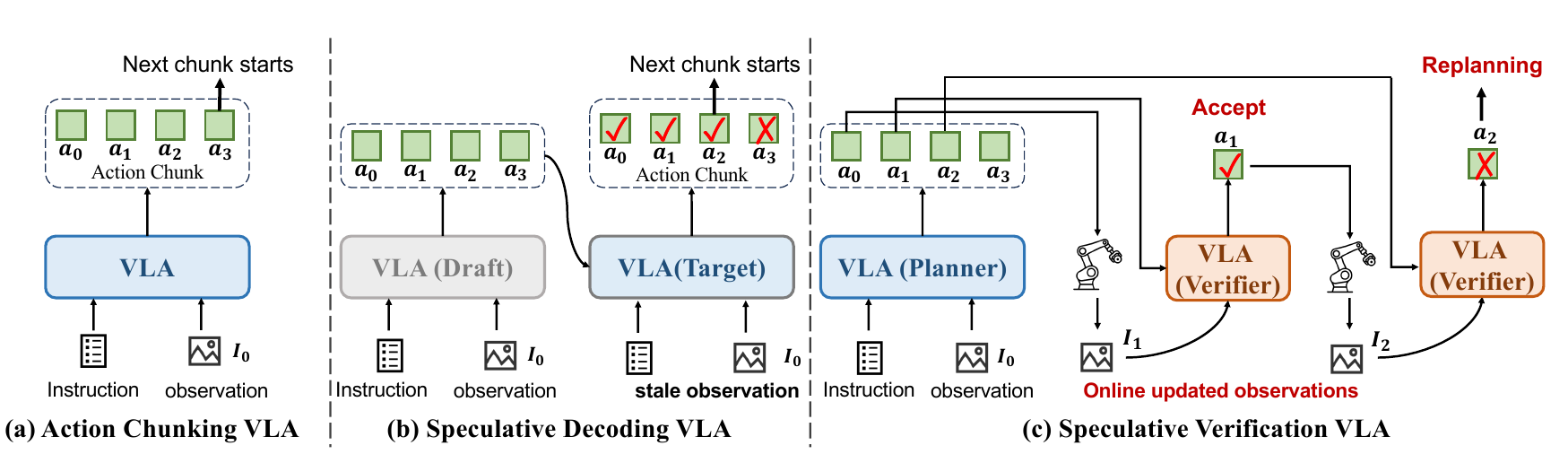}
  \caption{Comparison of action chunking, speculative decoding, and our proposed Speculative Verification VLA (SV-VLA).
(a) Action Chunking VLA predicts and executes an action chunk in an open-loop manner, so later actions may rely on stale observations.
(b) Speculative Decoding VLA employs a draft model to generate candidate actions, and a heavy target model to verify them in parallel. However, the verified chunk is still executed based on stale observations.
(c) Speculative Verification VLA (SV-VLA) performs chunk-level macro planning together with lightweight and frequent verification under online updated observations. During execution, it continuously verifies whether the current planned action remains valid under the latest observation and triggers replanning once a mismatch is detected. It combines the efficiency of chunk-level planning with the adaptability of closed-loop feedback, improving both responsiveness and robustness in dynamic environments.
}
\Description{Comparison among Action-Chunking VLA, Speculative Decoding VLA, and our proposed Speculative Verification VLA.}
  \label{fig:teaser}
\end{figure*}

Apart from action chunking mechanisms, recent works~\cite{wang2025spec} have begun to explore speculative decoding~\cite{chen2023accelerating, leviathan2023fast}, a technique originally developed for NLP, to improve the efficiency of action token generation in VLA control.
Specifically, as Figure~\ref{fig:teaser} (b) shows, speculative decoding accelerates action token generation by employing a lightweight draft model to propose candidate actions, which are then verified by the full VLA model.
However, when applied to VLA models, this paradigm remains limited by its open-loop assumption, since the verification of multiple candidate actions is conditioned on a fixed observation and cannot incorporate newly available observations during execution.
This is problematic for embodied decision-making, where accurate control requires conditioning actions on the latest observations in a closed-loop manner.

To address this limitation, we propose a \textbf{Speculative Verification} framework tailored for VLA models. 
Unlike recent speculative decoding-based methods~\cite{wang2025spec}, which perform action verification in an open-loop manner using a heavy verification model, our method employs a heavy draft model to predict multiple candidate action tokens in a single step and a lightweight verification model to sequentially validate them conditioned on updated observations.
As illustrated in Figure~\ref{fig:teaser} (c), this design enables predicting in an open-loop manner while verifying in a closed-loop manner, combining the efficiency of chunked prediction with the robustness of closed-loop control for fast yet reliable action generation.

More concretely, in \textbf{Speculative Verification}, at the beginning of each planning cycle, the heavy VLA generates a macro action chunk together with an internal feature that summarizes the high-level planning context.
During chunk execution, the heavy model is not called again unless replanning is needed.
Instead, a lightweight verifier operates at a higher frequency to continuously monitor the execution state.
The verifier combines current visual observations with the high-level feature from the heavy model to estimate a reference action that aligns with both the current state and the original task intent.
When the discrepancy between the reference action and the predicted action from the macro-planner exceeds a predefined threshold, the system treats the execution as deviating from the original plan and triggers replanning.
In this case, the remaining actions in the chunk are discarded, and the heavy VLA replans from the current state.
This design preserves the efficiency benefits of chunked control while restoring responsiveness to environmental changes.
Since the verifier is lightweight, it introduces negligible computational overhead.
As a result, the proposed \textbf{Deviation-based Replanning} mechanism preserves the efficiency of open-loop chunked prediction while restoring the responsiveness of closed-loop control.

In summary, our main contributions are as follows:
\begin{itemize}
    \item We introduce \textbf{Speculative Verification},
    a framework that leverages a full VLA model as a low-frequency macro-planner to generate action chunks, together with a lightweight verifier for high-frequency online verification based on real-time observations, effectively enabling open-loop prediction with closed-loop verification for efficient and reliable VLA control.
    \item We introduce a \textbf{Deviation-based Replanning} mechanism that detects execution mismatches and triggers replanning only when necessary, preserving the efficiency benefits of chunked control while restoring responsiveness to environmental changes.
    \item Extensive experiments on the LIBERO benchmark show that SV-VLA achieves a better balance between efficiency and robustness. Specifically, it improves the average success rate on the three subtasks by 11.4\% (from 79.5\% to 90.90\%), compared to the open-loop baseline.
\end{itemize}

\section{Related Works}
\subsection{Vision-Language-Action Models}
The integration of large-scale vision-language pretraining with robotics has catalyzed the development of Vision-Language-Action (VLA) foundation models~\cite{driess2023palm}. Early pioneering works, such as RT-1~\cite{brohan2022rt} and RT-2~\cite{zitkovich2023rt}, demonstrated that treating robot actions as textual tokens allows for the direct co-training of massive models on both internet-scale multimodal datasets and robotic demonstrations. Building upon this, recent open-weights models like OpenVLA~\cite{kim2024openvla}, Octo~\cite{team2024octo}, and $\pi_0$~\cite{black2024pi0} have achieved unprecedented zero-shot generalization and semantic reasoning in open-world manipulation tasks.
More recent efforts, such as Uni-Sight~\cite{li2025uni}, MoManipVLA~\cite{wu2025momanipvla}, and Matrix~\cite{chen2024expanding}, further improve VLA systems through unified multi-view alignment, richer multimodal fusion, and manipulation-aware action modeling, continuing the trend toward increasingly general and capable embodied foundation models. 
Recent works also begin to address long-horizon and interactive control, for example through unified hierarchical VLA control~\cite{yang2025lohovla, jing2025mixture,haon2025mechanistic} and real-time language-based correction~\cite{lynch2023interactive, yang2025efficientvla}. 
However, while these advances improve policy capability and task coverage, the large parameter scale of modern VLAs still imposes substantial inference latency, making it difficult to deploy them as high-frequency closed-loop controllers.

To mitigate the inference latency and compounding error problems in continuous control, Action Chunking has become a dominant paradigm in imitation learning and embodied AI. Originally popularized by Action Chunking with Transformers (ACT)~\cite{zhao2023learning} and Diffusion Policy~\cite{chi2025diffusion}, this approach predicts a multi-step sequence of future actions (a macro chunk) at each inference step. While executing a macro chunk open-loop ensures temporal smoothness and bridges the low-frequency inference gap, it sacrifices real-time reactivity. To recover closed-loop capabilities, some works propose receding horizon control or overlapping chunk aggregation~\cite{zhao2023learning, wu2025you, yang2026uaor}, but these methods still require querying the primary policy at high frequencies, which is costly for large VLAs. 

\subsection{Speculative Decoding}

Speculative Decoding (SD) has emerged as a cornerstone technique for accelerating autoregressive inference in large language models (LLMs)~\cite{leviathan2023fast, chen2023accelerating,li2024eagle,cai2024medusa}. 
In the standard SD paradigm, a lightweight draft model rapidly proposes candidate tokens, which are then verified in parallel by a larger target model. 
Medusa~\cite{cai2024medusa} is a representative approach that augments a single backbone LLM with multiple lightweight decoding heads, each trained to predict tokens at different future offsets.
EAGLE~\cite{li2024eagle} further improves speculative decoding by designing stronger and more efficient token extrapolation mechanisms from intermediate hidden representations.
Other extensions study tree-structured or multi-branch speculation, where multiple candidate continuations are proposed and verified jointly to increase parallelism and robustness~\cite{spector2023accelerating, miao2024specinfer}.
More broadly, recent efforts have investigated adaptive speculation lengths, hardware-aware scheduling, and system-level optimizations to better balance drafting overhead against verification efficiency~\cite{chen2023accelerating, jang2024lantern}.
This strategy is highly effective in NLP because the validity of future tokens depends only on the textual context, which is fully available once the candidate sequence has been drafted, without requiring any external feedback. 
However, extending SD to embodied control is fundamentally difficult: in robotics, future observations depend on the actual execution of prior actions and the subsequent evolution of the physical environment, so future actions cannot be verified in parallel in the same way. 
A few recent efforts, such as SpecVLA~\cite{wang2025spec}, have begun to explore speculative ideas for VLA inference, but these methods still mainly focus on inference acceleration under open-loop generation.

\section{Method}
\label{sec:method}

In this section, we present \textbf{SV-VLA}, a framework that combines low-frequency macro planning with high-frequency lightweight verification for efficient and adaptive Vision-Language-Action control.
We first introduce the standard action chunking formulation in VLA control(Sec.~\ref{subsec:preliminary}). 
We then analyze why standard speculative decoding cannot be directly applied to embodied control without sacrificing closed-loop responsiveness (Sec.~\ref{subsec:sd_mismatch}), which motivates our design. 
Based on this observation, we present the decoupled architecture of SV-VLA, consisting of a heavy {Macro-Planner} for low-frequency action chunk generation and a lightweight action verifier for high-frequency verification (Sec.~\ref{subsec:actguard_arch}). 
Finally, we formalize the Deviation-based Replanning mechanism (Sec.~\ref{subsec:adaptive_rule}) and the training strategy for the verifier module (Sec.~\ref{subsec:training}).

\subsection{Preliminary}
\subsubsection{Action Chunking VLA}
\label{subsec:preliminary}
In embodied manipulation, a Vision-Language-Action (VLA) model aims to learn a policy $\pi$ that maps a visual observation $I_t$, a natural language instruction $L$, and a proprioceptive state $s_t$ to continuous robot actions.
To mitigate the inference latency of Large Vision-Language Models (LVLMs) and bridge the gap with high-frequency control, recent VLA systems such as OpenVLA-OFT adopt \textit{action chunking}.
Instead of predicting a single action, the model generates a sequence of $K$ future actions at step $t$:
\begin{equation}
    A_t = [a_t, a_{t+1}, \dots, a_{t+K-1}] = \pi_{\theta}(I_t, L, s_t),
\end{equation}
where $\pi_{\theta}$ denotes the heavy VLA policy and $K$ is the chunk size.

Action chunking amortizes the expensive inference cost of the LVLM over multiple control steps and often improves motion smoothness.
However, the predicted chunk is executed in an open-loop manner between two planning boundaries.
As the execution proceeds, compounding actuation errors, object perturbations, and external disturbances may gradually invalidate the assumptions under which the chunk was generated.
This effect becomes particularly severe when $K$ is large (e.g., $K=64$): while long chunks improve computational efficiency, they also increase exposure to temporal drift and error accumulation in dynamic environments.

A straightforward solution is to invoke the heavy VLA for closed-loop replanning at every step.
However, because the per-call latency of the LVLM is often comparable to or larger than the control interval, such dense replanning is typically impractical for real-time manipulation.

\begin{figure*}
  \includegraphics[width=0.95\textwidth]{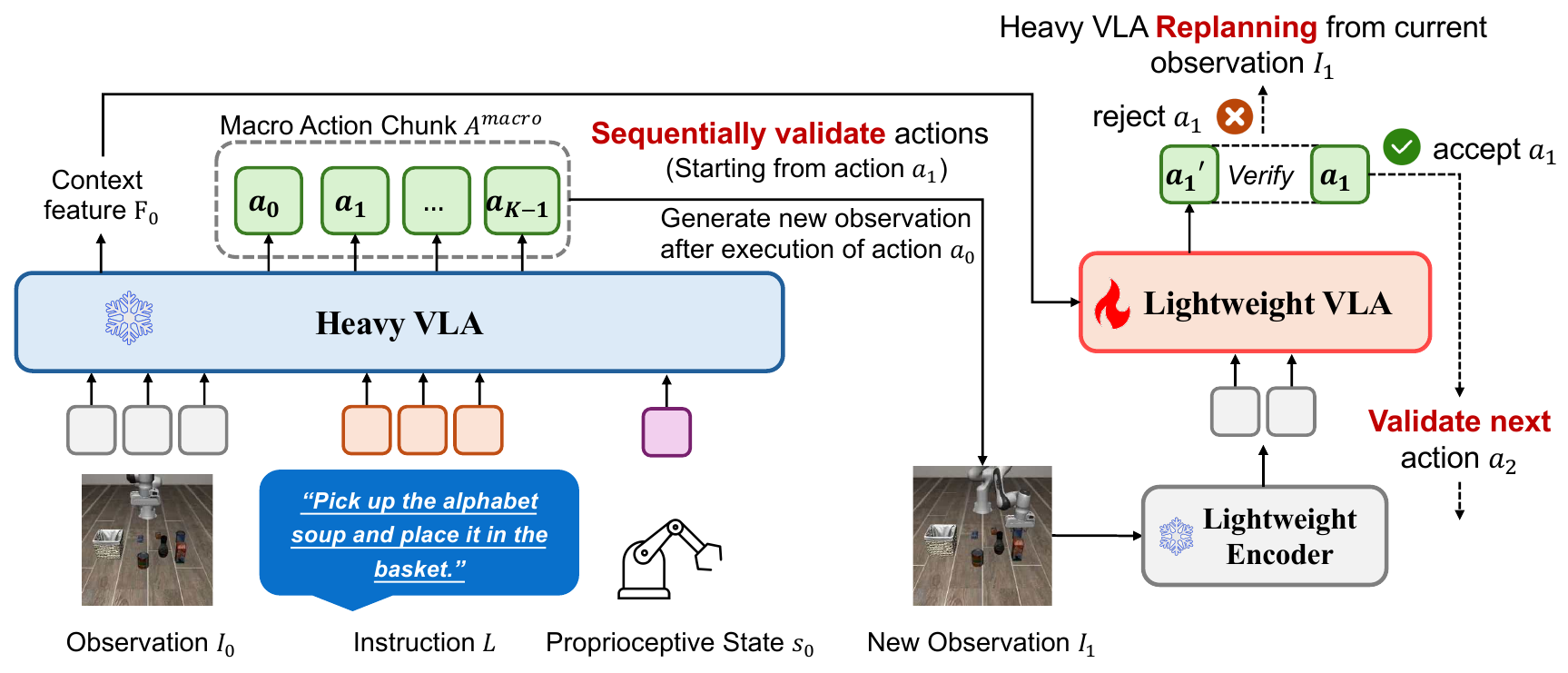}
  \caption{
Overview of SV-VLA. At each planning boundary $T_0$, a frozen heavy VLA takes the current observation $I_0$, language instruction $L$, and proprioceptive state $s_0$ as input, and outputs a macro action chunk $A^{macro}$ together with a planning context feature $F_0$. 
During execution, a lightweight verifier runs at control frequency. Given the latest observation $I_1$ and the planning context feature $F_0$, it predicts a reference action $a_1'$, which is compared with the current planned action $a_1$ from the macro chunk. If the discrepancy is below a threshold, the planned action $a_1$ is accepted and executed; otherwise, it is rejected, the remaining chunk is discarded, and the heavy VLA replans from the current state.}
\Description{Overview of SV-VLA.}
  \label{fig:archi}
\end{figure*}

\subsubsection{From Speculative Decoding to Speculative Verification}
\label{subsec:sd_mismatch}
We next revisit why standard speculative decoding (SD) is ill-suited to embodied control, and how this motivates our design.
In autoregressive language modeling, SD accelerates generation by using a lightweight draft model to propose multiple future tokens and a heavy target model to verify them in parallel~\cite{leviathan2023fast}.
This parallelism is possible because token verification depends only on the already available symbolic context.

Formally, given a heavy target model $\pi_{\text{target}}$ and a lightweight draft model $\pi_{\text{draft}}$, SD first generates a candidate sequence $\hat{Y} = [\hat{y}_1, \dots, \hat{y}_K]$ autoregressively:
\begin{equation}
    \hat{y}_t \sim \pi_{\text{draft}}(\cdot \mid Y_{<t}), \quad t \in \{1, \dots, K\},
\end{equation}
where $Y_{<t}$ denotes the previously accepted prefix.
The heavy model can then score all drafted tokens in parallel, because the drafted sequence itself provides the future context needed for verification.
No interaction with an external environment is required.

This assumption does not hold in embodied manipulation.
In robotics, the validity of a future action depends on future observations, which are not available in advance but are induced by executed actions and environment dynamics.
Let $\mathcal{M}$ denote the environment transition model.
Then the next observation is given by
\begin{equation}
    I_{t+1} \sim \mathcal{M}(\cdot \mid I_t, a_t, s_t).
\end{equation}
Therefore, whether an action $a_{t+1}$ remains valid cannot be determined solely from the current observation $I_t$; it depends on the actual next observation $I_{t+1}$, which only becomes available after executing $a_t$.
As a result, action verification in embodied control cannot be parallelized in the same way as token verification in language generation.
Instead, it must be interleaved with environment interaction and performed sequentially over time.

This difference has a crucial latency implication.
Let $\mathcal{T}_{\text{heavy}}$ denote the inference latency of the heavy VLA and $\mathcal{T}_{\text{ctrl}}$ the control interval.
For real-time deployment, this requires $\mathcal{T}_{\text{heavy}} \leq \mathcal{T}_{\text{ctrl}}$.
In practice, however, large VLA models often have inference latency comparable to or exceeding the control interval, i.e., $\mathcal{T}_{\text{heavy}} \gtrsim \mathcal{T}_{\text{ctrl}}$.
Therefore, SD-style designs that use the heavy VLA as an online verifier under updated observations are typically impractical for real-time manipulation.

As a practical workaround, prior SD-style VLA methods~\cite{wang2025spec} typically verify drafted actions using only the current, stale observation, rather than the future observations that would actually arise during execution.
While this avoids repeated heavy-model inference, it introduces a fundamental mismatch between verification and the true execution trajectory. This analysis suggests that efficient VLA control requires a different design from standard speculative decoding: action chunks should still be generated by a strong planner to preserve long-horizon competence, but verification must be performed sequentially under updated observations using a lightweight module that can run at control frequency. SV-VLA is built exactly on this principle.

\subsection{Speculative Verification VLA (SV-VLA)}

\subsubsection{Open-Loop Planning and Closed-Loop Verification}
\label{subsec:actguard_arch}
These observation directly motivates SV-VLA's central design principle: embodied control should \emph{plan expensively but infrequently, and verify cheaply but continuously}.
As Figure~\ref{fig:archi} shows, at each planning boundary $T_0$ (either at the beginning of a task or immediately after a replanning event), we invoke the heavy VLA once to produce both a macro action chunk and a compact representation of the planning context.
During both deployment and the specific verifier training phase, the weights of the heavy VLA remain frozen.

Given the current visual observation $I_0$, language instruction $L$, and proprioceptive state $s_0$, the heavy VLA outputs:
\begin{equation}
    (A^{macro}, F_0) = \pi_{\theta}(I_0, L, s_0),
\end{equation}
where $A^{macro} = [a_0, a_1, \dots, a_{K-1}]$ is the planned macro action chunk for subsequent execution; and $F_0$ is a contextual reference feature extracted from the second-to-top Transformer layer of the heavy VLA, summarizing the multimodal intent and scene context at the planning boundary.

The generated chunk $A^{macro}$ can be interpreted as a speculative open-loop prediction of future actions under the assumption that the environment evolves as anticipated from the state at $T_0$.
The first action $a_0$ is executed immediately, causing the environment to transition to a new physical state and generating a new observation at $T_1$.
At any subsequent step $t \in \{1,2,\dots,K-1\}$, instead of either blindly executing the pre-planned action $a_t$ or re-invoking the heavy VLA, a lightweight verifier module assesses whether $a_t$ remains valid under the current observation.

At step $t$, the updated visual observation $I_t$ is processed by a frozen lightweight vision backbone $\phi_{vit}$ (e.g., ViT-Tiny) to produce a visual feature:
\begin{equation}
    E_t = \phi_{vit}(I_t).
\end{equation}
To keep the verifier grounded in both the current observation and the original plan, we fuse the visual feature $E_t$ with the planning feature $F_0$:
\begin{equation}
    Z_t = \mathrm{FC}\big([E_t \parallel F_0]\big),
\end{equation}
where $\parallel$ denotes feature concatenation. The fused representation is then passed to a lightweight verification head to predict a reference action
\begin{equation}
    a'_t = \pi_{verify}(Z_t).
\end{equation}
This action is not executed directly. Instead, it serves as an online reference for checking whether the current macro action $a_t$ is still locally valid.

This yields the following role decomposition:
\begin{itemize}
    \item \textbf{Heavy VLA Model as Open-Loop Low-Frequency Macro-Planner.} The heavy VLA is invoked sparsely, only at planning boundaries, to generate a long-horizon speculative action chunk and the associated contextual feature.
    \item \textbf{Lightweight Module as Closed-Loop High-Frequency Verifier.} The verifier operates at each control step using the latest observation and the frozen planning context.
\end{itemize}

\subsubsection{Deviation-based Replanning}
\label{subsec:adaptive_rule}

At each execution step $t$, the verifier compares the pre-planned macro action $a_t \in A^{macro}$ with its closed-loop verification action $a'_t$ under the current observation.
We quantify their discrepancy using the L1 distance:
\begin{equation}
    \mathcal{E}_t =\text{norm}( \|a'_t - a_t\|_1),
\end{equation}
where norm($\cdot$) rescales the raw discrepancy to $[0,1]$ according to the valid range of the action space.
Intuitively, $\mathcal{E}_t$ measures how far the original macro action has drifted from an action that is locally consistent with the current visual observation.

Given a safety threshold $\tau$ ($\tau \in (0,1)$), we apply the following binary execution rule:
\begin{equation}
    \pi_{exec}(t)=
    \begin{cases}
      a_t, & \text{if } \mathcal{E}_t \leq \tau,\\[6pt]
      \dot a_t, \quad \text{where } [\dot a_t, \dots] = \pi_{\theta}(I_t, L, s_t), & \text{if } \mathcal{E}_t > \tau,
   \end{cases}
\end{equation}
where $\dot a_t$ denotes the first action of a newly replanned macro-action sequence generated by the heavy VLA model from the current timestep $t$.
When $\mathcal{E}_t \leq \tau$, the current state is still considered compatible with the assumptions under which $A^{macro}$ was generated, so the robot continues executing the original planned action $a_t$. In contrast, when $\mathcal{E}_t > \tau$, the discrepancy suggests that execution has drifted sufficiently far from the original open-loop plan. In this case, SV-VLA discards the remaining suffix $[a_t, \dots, a_{K-1}]$ and invokes the heavy VLA to replan from the current state $(I_t, s_t)$, which incurs one additional heavy-model call with latency $\mathcal{T}_{heavy}$.
Note that the verifier does not replace the macro action; it only assesses whether the current execution remains consistent with the original plan.

A key advantage of this design is that verification and replanning operate at different timescales. Since the visual backbone is lightweight and the verification head is shallow, the verifier can run at or near the control frequency, i.e., $\mathcal{T}_{verify} \lesssim \mathcal{T}_{ctrl}$. By contrast, the heavy VLA acts as a low-frequency planner with latency $\mathcal{T}_{heavy} \gtrsim \mathcal{T}_{ctrl}$. This decoupling restores closed-loop responsiveness while preserving the long-horizon planning capability of the heavy model.

This mechanism also yields a simple efficiency characterization between two extreme cases. In the best case, the verification result is always correct and every macro action chunk is fully executed, so one heavy VLA call is amortized over $K$ control steps. The per-step inference cost then approaches 
$    \mathcal{C}_{step}^{\min}=
    \frac{\mathcal{C}_{VLA}}{K}+
    \mathcal{C}_{verify}$
 , which is close to standard open-loop chunking.
 In the worst case, replanning is triggered at every step, so the system degenerated to step-wise execution. Since the verifier is lightweight and its inference cost is negligible compared with that of the full VLA model, the per-step cost in this case becomes $  \mathcal{C}_{step}^{\max}=\mathcal{C}_{VLA}+\mathcal{C}_{verify}\approx \mathcal{C}_{VLA}.$
 Therefore, SV-VLA naturally operates between these two bounds: it approaches the efficiency of chunked open-loop control when prediction remains reliable, while retaining the ability to fall back to frequent replanning when long-horizon execution becomes unsafe.

\subsection{Training Strategy}
\label{subsec:training}

The training objective of the verifier is to learn a lightweight real-time reference policy that can quickly assess whether the current macro action remains locally valid under the latest observation.
During the training phase, the heavy VLA $\pi_{\theta}$ remains frozen and serves only as a provider of the macro chunk $A^{macro}$ and the contextual reference feature $F_0$.

For each subsequent timestep $t \in \{1, \dots, K-1\}$, the verifier takes the current observation $I_t$ together with the frozen planning feature $F_0$ as input and predicts a reference action. 
The verifier is trained with an L1 regression loss:
\begin{equation}
    \mathcal{L}_{verify}
    = \frac{1}{K-1}\sum_{t=1}^{K-1} \|a'_t - \hat{a}_t\|_1.
\end{equation}
where $\hat{a}_t$ is the ground-truth action at timestep $t$.
In this way, the verifier learns to produce reference actions conditioned on both the original high-level plan and the latest observation for online verification.

By freezing the heavy VLA and training only the lightweight verifier, SV-VLA preserves compatibility with existing pretrained VLA models while introducing minimal additional inference overhead.
This design allows the expensive planner to remain unchanged, and concentrates all adaptation capacity on the module responsible for real-time closed-loop verification.

\begin{algorithm}[htbp]
\caption{Inference Pipeline of SV-VLA}
\label{alg:actguard_inference}
\textbf{Require:} Heavy VLA model $\pi_{vla}$, Lightweight verifier $\pi_{verify}$, Efficient Vision Encoder $\phi_{vit}$ \\
\textbf{Require:} Language instruction $L$, Max chunk size $K$, Safety threshold $\tau$
\begin{algorithmic}[1]
\State Obtain Initial observation $I_{curr}$ and state $s_{curr}$
\While{task is not completed}
    \State \textcolor{gray}{\% --- 1. Macro Planning \& Reference Extraction ($T_0$) ---}
    \State $I_0 \gets I_{curr}$, $s_0 \gets s_{curr}$
    \State $A^{macro}, F_0 \gets \pi_{vla}(I_0, L, s_0)$ \Comment{Macro Planning chunk and extract context feature}
    \State Execute $a_0 \in A^{macro}$
    \State Observe next state $(I_1, s_1)$
    
    \State \textcolor{gray}{\% --- 2. High-Frequency Lightweight Verification ($T_t$) ---}
    \For{$t = 1, 2, \dots, K-1$}
        \State $I_t \gets I_{curr}$
        \State $E_t \gets \phi_{vit}(I_t)$ \Comment{Extract lightweight visual feature}
        
        \State \textcolor{gray}{\% Verifier Estimation}
        \State $Z_t \gets \operatorname{FC} \big( [E_t \parallel F_0] \big)$ \Comment{Feature Fusion}
        \State $a'_t \gets \operatorname{Attention}(Z_t)$ \Comment{Prediction by Verifier}
        
        \State \textcolor{gray}{\%Deviation-based Replanning}
        \State $\mathcal{E}_t \gets \text{norm}( \| a'_t - a_t \|_1)$ \Comment{Discrepancy against pre-planned $a_t$}
        
        \If{$\mathcal{E}_t > \tau$}
            \State \textbf{break} \Comment{Replan Triggered: Abort rest of $A^{macro}$}
        \Else
            \State Execute $a_t \in A^{macro}$ \Comment{Accept the Planned Actions}
            \State Observe next state $(I_{curr}, s_{curr})$
        \EndIf
    \EndFor
    \State \textcolor{gray}{\% Chunk completed naturally or prematurely aborted; loop restarts for heavy replanning.}
\EndWhile
\end{algorithmic}
\end{algorithm}


\section{Experiments}
\subsection{Datasets}
We use the LIBERO benchmark~\cite{liu2023libero} in our experiments. 
LIBERO is designed to evaluate robot manipulation and policy generalization across diverse tasks and environments, providing a standardized simulation benchmark for embodied AI. 
It consists of three different task suites: LIBERO-Goal, LIBERO-Spatial, and LIBERO-Object, which cover different sources of variation, including goals, spatial relations, and object categories. 
To evaluate the effectiveness of our method, we conduct online rollout experiments in simulation and report the task success rate and the inference latency.

\subsection{Implement Details}
We adopt the OpenVLA-OFT~\cite{oft} with chunk size 64 as our heavy VLA model.
Training is conducted on 4 NVIDIA A100 GPUs, and all validation experiments are performed on a single NVIDIA V100 GPU to ensure a fair comparison. 
We use AdamW for optimization with a learning rate of $8\times10^{-4}$ and a batch size of 64. 
During evaluation, we perform online rollouts in the LIBERO simulator with batch size of 1 and report the average task success rate and inference speed.

\subsection{Main Results}
To evaluate the trade-off between inference efficiency and execution reliability, we compare different methods under varying macro-action chunking strategies on the LIBERO benchmark. 
A key challenge in action chunking is that larger chunks reduce the frequency of expensive heavy-VLA inference, but also make execution more vulnerable to stale observations, temporal drift, and compounding errors. 
Table~\ref{tab:main_performance} shows that SV-VLA achieves a favorable balance between these two objectives. 
Overall, our method is consistently much faster than the short-chunk baseline with $K=8$, while substantially more reliable than the long-chunk open-loop baseline with $K=64$. 
Using the same long chunk size of $K=64$, SV-VLA improves the average success rate from 79.5\% to 90.90\%, while still maintaining a 2.17$\times$ speed-up over the $K=8$ baseline. 
Across individual task suites, the gains are also consistent: on LIBERO-Goal, SV-VLA improves over open-loop $K=64$ from 93.2\% to 94.4\%; on LIBERO-Object, from 77.2\% to 95.3\%; and on LIBERO-Spatial, from 68.0\% to 83.0\%. 
This result suggests that lightweight online verification can recover much of the robustness lost in long-horizon open-loop execution, without sacrificing most of its computational advantage.

The comparison between the two open-loop baselines highlights the fundamental trade-off of naive action chunking. 
Using a short chunk size ($K=8$) yields the strongest performance, achieving an average success rate of 96.0\%, but requires the most frequent heavy-VLA inference. 
In contrast, increasing the chunk size to $K=64$ improves inference speed to 3.15$\times$ on average, but causes the average success rate to drop sharply to 79.5\%. 
This degradation is particularly severe on more challenging suites such as LIBERO-Object and LIBERO-Spatial, where success falls from 98.3\% to 77.2\% and from 93.0\% to 68.0\%, respectively. 
These results confirm that efficiency cannot be obtained by simply extending open-loop execution alone.

We further compare against Speculative Decoding, which also aims to accelerate inference by introducing an auxiliary smaller model. 
Overall, SV-VLA achieves a better trade-off between speed and control performance. 
Speculative decoding attains 81.7\% average success with a 1.36$\times$ speed-up, whereas SV-VLA reaches 90.90\% average success with a substantially larger 2.17$\times$ speed-up. 
On LIBERO-Goal and LIBERO-Object, SV-VLA consistently outperforms speculative decoding by large margins (94.4\% vs. 74.4\%, and 95.3\% vs. 85.0\%, respectively). 
On LIBERO-Spatial, speculative decoding obtains slightly higher success (85.8\% vs. 83.0\%), but this comes with much lower acceleration than SV-VLA (1.28$\times$ vs. 2.49$\times$). 
These results suggest that, while speculative decoding can provide limited inference benefits, simply accelerating action generation under stale context does not reliably resolve the execution mismatch in embodied control. 
In contrast, SV-VLA performs online verification under updated observations and triggers replanning only when necessary, resulting in stronger balance of efficiency and robustness.

\begin{table}[htbp]
  \caption{Main results on the LIBERO benchmark. We compare different methods on three task suites. We report success rate and inference speed-up ratios relative to the open-loop baseline with chunk size $K=8$. \underline{\textbf{Bold}} indicates the best result, and \underline{underline} indicates the second-best result. Overall, SV-VLA achieves a better balance between efficiency and robustness.}
  \label{tab:main_performance}
  \resizebox{\columnwidth}{!}{
  \begin{tabular}{lccc}
    \toprule
    Method & Chunk Size & Success (\%)$\uparrow$ & Speed $\uparrow$\\
    \midrule

    \multicolumn{4}{c}{\textit{LIBERO-Goal}} \\
    \midrule
    BASE ($K=8$~\cite{oft}) & $8$  & \underline{\textbf{96.8}} & 1.00 \\
    BASE ($K=64$)                   & $64$ & 93.2 & \underline{\textbf{2.78}} \\
    Speculative Decoding~\cite{wang2025spec} & $4$ & 74.4 & 1.42 \\
    Speculative Verification (Ours) & $64$ & \underline{94.4} & \underline{1.56} \\
    \midrule

    \multicolumn{4}{c}{\textit{LIBERO-Object}} \\
    \midrule
       BASE ($K=8$~\cite{oft}) & $8$  & \underline{\textbf{98.3}} & 1.00 \\
     BASE ($K=64$)                 & $64$ & 77.2 & \underline{\textbf{3.42}} \\
    Speculative Decoding~\cite{wang2025spec} & $4$ & 85.0 & 1.38 \\
    Speculative Verification (Ours) & $64$ & \underline{95.3} & \underline{2.46} \\
    \midrule

    \multicolumn{4}{c}{\textit{LIBERO-Spatial}} \\
    \midrule
   BASE ($K=8$~\cite{oft}) & $8$  & \underline{\textbf{93.0}} & 1.00 \\
     BASE ($K=64$)               & $64$ & 68.0 & \underline{\textbf{3.22}} \\
    Speculative Decoding~\cite{wang2025spec} & $4$ & \underline{85.8} & 1.28 \\
    Speculative Verification (Ours) & $64$ & 83.0 & \underline{2.49} \\
    \midrule

    \multicolumn{4}{c}{\textit{Average}} \\
    \midrule
   BASE ($K=8$~\cite{oft}) & $8$  & \underline{\textbf{96.0}} & 1.00 \\
     BASE ($K=64$)                  & $64$ & 79.5 & \underline{\textbf{3.15}} \\
    Speculative Decoding~\cite{wang2025spec} & $4$ & 81.7 & 1.36 \\
    Speculative Verification (Ours) & $64$ & \underline{90.9} & \underline{2.17}  \\
    
    \bottomrule
  \end{tabular}
  }
\end{table}

\begin{table}[t]
  \centering
  \caption{Inference time breakdown.
  We compare BASE ($K=8$ and $K=64$) and SV-VLA in terms of average macro-planner/verifier calls per episode, average inference time per episode, together with success rate. \underline{\textbf{Bold}} indicates the best result, and \underline{underline} indicates the second-best result. Compared with BASE ($K=64$), SV-VLA improves success through online verification and replanning.
}
  \label{tab:inference_analysis}
  \resizebox{\columnwidth}{!}{
  \begin{tabular}{lrrr}
    \toprule
    Metric & BASE ($K=8$) & BASE ($K=64$) & SV-VLA ($K=64$)\\
    \midrule
    Macro-Planner calls / episode & 14.5 & 4.2 & 6.7 \\
    Verifier calls/ episode      & 0    & 0   & 13.3 \\
    Average time / episode       & 15.9\,s &\underline{\textbf{5.7\,s}} &\underline{8.8\,s} \\
    Success rate             & \underline{\textbf{96.00\%}} & 79.47\% & \underline{90.90\%} \\
    \bottomrule
  \end{tabular}
  }
\end{table}

To better understand where the efficiency gain comes from, we further analyze the inference-time composition of SV-VLA in Table~\ref{tab:inference_analysis}. 
The key observation is that the two components of our system operate at very different cost scales: each macro-planner call takes 1.373\,s on average, whereas each verifier call takes only 0.081\,s. 
This large cost gap directly motivates our design of invoking the heavy macro-planner sparsely while relying on frequent lightweight verification for online monitoring.

The call statistics in Table~\ref{tab:inference_analysis} further explain the runtime trade-off achieved by SV-VLA. 
Compared with BASE ($K=8$), which requires 14.5 macro-planner calls per episode, SV-VLA reduces the number of expensive macro-planner calls to 6.7 by executing longer chunks and replanning only when necessary. 
Although SV-VLA additionally performs 13.3 verifier calls per episode, these checks are much cheaper than heavy-VLA inference, so the total inference time is reduced from 15.9\,s to 8.8\,s per episode. 
Compared with BASE ($K=64$), SV-VLA incurs only a moderate increase in inference time (8.8\,s vs.\ 5.7\,s), because it performs online verification and occasional replanning instead of blindly executing the full chunk open-loop. 
This modest overhead, however, yields a substantial gain in control reliability, improving the success rate from 79.47\% to 90.90\%. 
Overall, these results show that SV-VLA preserves much of the efficiency benefit of long-horizon chunking while recovering a substantial portion of the robustness lost under naive open-loop execution, thereby striking a practical balance between dense heavy-VLA replanning and blind long-horizon open-loop control.

\subsection{Ablations}

\begin{table}[t]
\centering
\caption{Ablation study of SV-VLA. We report success rate (Success), inference speed-up ratios (Speed), and the average number of executed action steps before each replanning event (Steps) on Libero-Spatial. SV-VLA with $K=64$ uses the default replanning threshold $\tau=0.2$. \underline{\textbf{Bold}} indicates the best result, and \underline{underline} indicates the second-best result.}
\label{tab:ablation_main}
\resizebox{0.48\textwidth}{!}{
\begin{tabular}{lccc}
\toprule
Variant & Success (\%) $\uparrow$ & Speed $\uparrow$ & Steps \\
\midrule
\multicolumn{4}{c}{\textit{Baselines}} \\\hline
BASE ($K=8$)              & 93.0 & 1.00 & 8 \\
BASE ($K=64$)              & 68.0 & 3.22 & 64 \\
\midrule
\multicolumn{4}{c}{\textit{SV-VLA Variants}} \\\hline
w/o context feature $F_0$      & \underline{73.7}& 1.46 & 24.9 \\
w/o current observation $I_t$ & 63.7 & 1.19 & 13.1 \\
w/o replanning & 15.5 & \underline{\textbf{3.13}} & 64 \\
SV-VLA ($K=64, \tau=0.2$)       & \underline{\textbf{83.0}} & \underline{2.49} & 37.2 \\
\midrule
\multicolumn{4}{c}{\textit{Sensitivity of Threshold $\tau$ at $K=64$}} \\\hline
SV-VLA ($\tau=0.1$)                  & \underline{\textbf{83.1}} & 2.17 & 28.0 \\
SV-VLA ($\tau=0.2$)               & \underline{83.0} & \underline{2.49} & 37.2 \\
SV-VLA ($\tau=0.4$)              & 77.4 & \underline{\textbf{2.60}} & 40.5 \\
\bottomrule
\end{tabular}
}
\end{table}

\begin{figure*}[htbp]
  \includegraphics[width=0.95\textwidth]{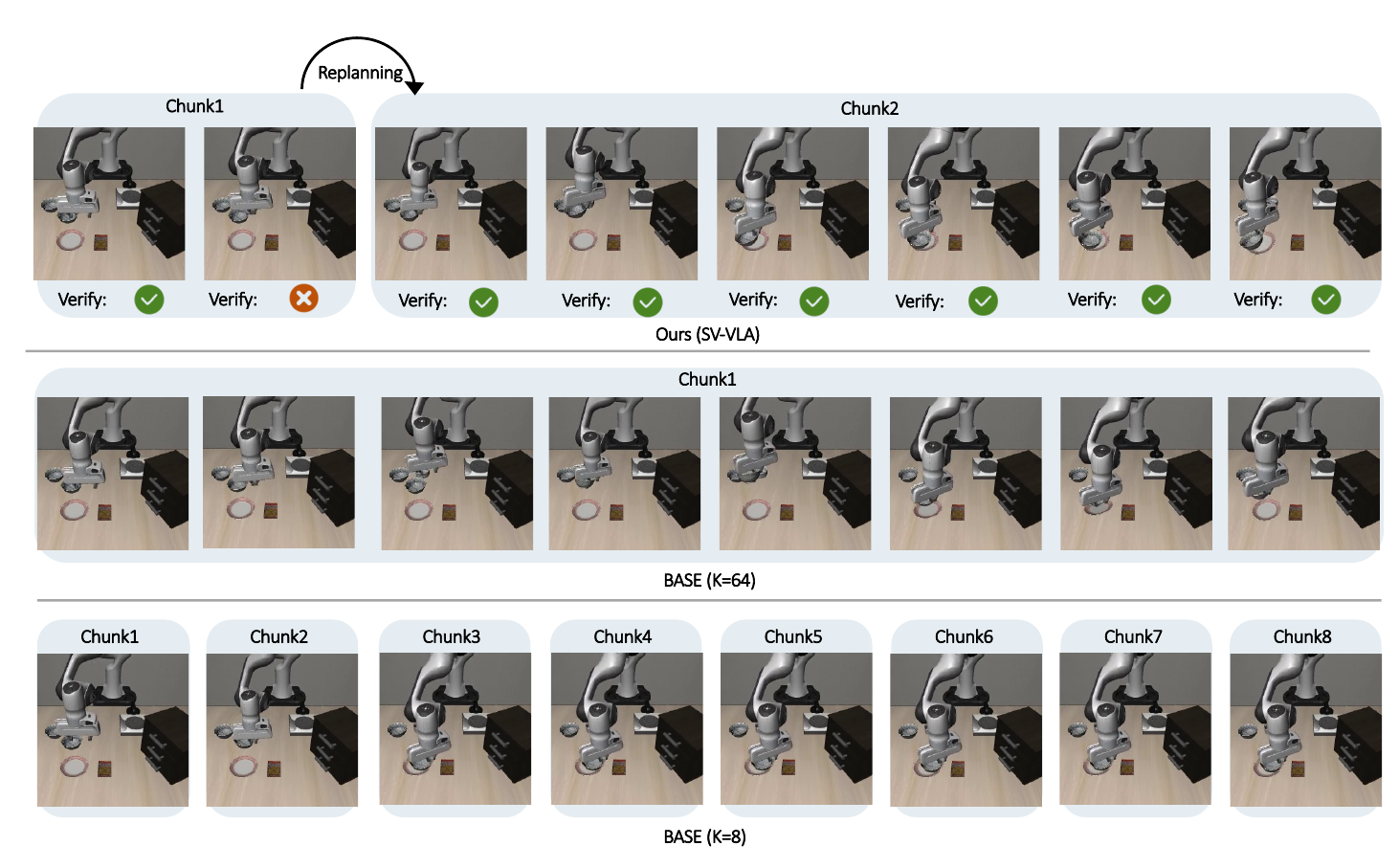}
\caption{Qualitative comparison on the task: ``pick up the black bowl between the plate and the ramekin and place it on the plate.'' 
SV-VLA detects during execution that the bowl has not been successfully grasped, interrupts the current macro action chunk, and replans from the latest observation, which enables successful task completion. 
In contrast, the open-loop baseline with $K=64$ executes the entire action chunk without correction and fails to recover from the grasping error, while the $K=8$ baseline remains robust through frequent replanning at a much higher inference cost.}

  \label{fig:vis}
\end{figure*}

Table~\ref{tab:ablation_main} presents the ablation results of SV-VLA. 
\textbf{BASE ($K=8$)} and \textbf{BASE ($K=64$)} denote the original action-chunking VLA baseline that executes a predicted macro action chunk of horizon $K$ in an open-loop manner, without online verification or replanning. 
\textbf{w/o context feature $F_0$} removes the planning context feature from the verifier and performs online verification using only the current observation.
\textbf{w/o current observation $I_t$} removes the current visual observation from the verifier and uses only the frozen planning context feature for action verification.
\textbf{w/o replanning} removes the deviation-based replanning mechanism and directly executes the verifier-predicted action at each step, without invoking the heavy VLA again after the initial planning step. 
It can be viewed as a \textit{speculative decoding} variant in the embodied setting, where verification is conditioned on a fixed planning representation rather than updated observations during execution. 
Finally, \textbf{SV-VLA ($K=64, \tau=0.2$)} denotes our proposed model, and we further study the sensitivity to the replanning threshold $\tau$.

We first examine the two inputs to the verifier. 
Removing the planning context feature (\textbf{w/o context feature $F_0$}) reduces success from 83.0\% to 73.7\%, indicating that the high-level planning context provides useful guidance for maintaining consistency with the original macro plan. 
Removing the current observation (\textbf{w/o current observation $I_t$}) causes a larger drop, to 63.7\%. 
This result shows that access to real-time visual feedback is even more critical: without the latest observation, the verifier cannot reliably detect execution drift or scene changes, and thus cannot perform truly closed-loop verification. 
Together, these results suggest that effective verification requires jointly reasoning over both the global planning context and the current execution state.

We next analyze the role of replanning. 
The \textbf{w/o replanning} variant performs dramatically worse than all other configurations, achieving only 15.5\% success despite its high speed. 
This confirms that lightweight verification alone is insufficient once the planned trajectory becomes invalid. 
The key benefit of SV-VLA is therefore not to replace the heavy VLA with a faster verifier, but to use the verifier to monitor execution and trigger replanning when necessary. 
Without this recovery mechanism, execution errors quickly accumulate and cannot be corrected.

We also observe the expected threshold trend: a smaller $\tau$ makes the system more conservative and increases replanning frequency, whereas a larger $\tau$ allows longer open-loop execution at the cost of reduced robustness.
In practice, this provides a simple task-level control knob: $\tau$ can be selected according to the robustness demands of a given task to better balance execution reliability and inference efficiency.

\subsection{Qualitative Results}
Figure~\ref{fig:vis} provides a qualitative comparison of SV-VLA with standard action chunking VLA models. 
In the top row, SV-VLA first generates a macro action chunk and begins execution while the verifier continuously monitors the latest observations. 
During the execution of the first chunk, the verifier detects that the bowl has not been successfully grasped and that the current trajectory is no longer consistent with the original plan. 
As a result, the remaining actions in the chunk are discarded and the heavy VLA is invoked to replan from the current state. 
With the updated plan, the robot successfully re-attempts the manipulation and completes the task. 
This example illustrates the key advantage of SV-VLA: it can detect execution failures early and recover in time, while still preserving the efficiency benefits of long-horizon chunking.

The middle row shows {BASE ($K=64$)}, where a single long action chunk is executed in an open-loop manner. 
Although this strategy is computationally efficient, it lacks any mechanism to detect execution failures after the chunk has been issued. 
As a result, once the grasp does not proceed as expected, the execution continues without correction, and the deviation accumulates over time.
In contrast, the bottom row shows {BASE ($K=8$)}. 
Because replanning happens every 8 actions, execution remains stable and closely follows the desired trajectory. 
However, each chunk requires a new call to the heavy VLA to generate the next action chunk, leading to frequent large-model inference and substantially higher computational cost.


\section{Conclusion}

We introduced SV-VLA, a framework for open-loop planning and closed-loop verification for action chunking VLA. 
Motivated by the mismatch between standard speculative decoding and embodied control, SV-VLA adopts a Speculative Verification design: the heavy VLA is used as a low-frequency macro-planner, while a lightweight verifier verifies execution at high frequency using real-time observations. 
This enables Deviation-based Replanning, allowing the system to execute only the valid prefix of a planned action chunk and replan when the current state no longer matches the original plan.
Our results show that this design improves manipulation robustness  while significantly reducing the amortized cost relative to dense closed-loop replanning. 
SV-VLA is also lightweight to adapt, since both the heavy VLA and the visual encoder remain frozen and only the verifier is trained.

\paragraph{Limitations and Future Work.}
The current system relies on a fixed discrepancy threshold $\tau$, which may not be optimal across heterogeneous tasks. 
Future work could explore adaptive triggering strategies and extend the verifier to provide small local corrections before falling back to heavy-model replanning.

\bibliographystyle{ACM-Reference-Format}
\bibliography{sample-base}


\vspace*{0pt}
\end{document}